\documentclass{article}

\usepackage{arxiv}

\usepackage[utf8]{inputenc} 
\usepackage[T1]{fontenc}    
\usepackage{hyperref}      
\usepackage{url}            
\usepackage{booktabs}       
\usepackage{amsfonts}       
\usepackage{nicefrac}       
\usepackage{microtype}      
\usepackage{lipsum}
\usepackage{graphicx}
\usepackage{algorithmic}
\usepackage{algorithm}
\usepackage{float}
\usepackage{titlesec}
\usepackage{subcaption}

\usepackage{amsmath}

\graphicspath{ {./images/} }

\title{Transferring Spatial Filters via Tangent Space
Alignment in Motor Imagery BCIs}

\author{
\textbf{Tekin Gunasar$^1$, Virginia de Sa$^{1,2}$} \\
$^1$Cognitive Science, $^2$Halicioglu Data Science Institute \\
UC San Diego \\
La Jolla, CA, USA \\
\texttt{tgunasar@ucsd.edu}
}

\providecommand{\keywords}[1]{\textbf{\textit{Keywords---}} #1}

\begin{document}

\maketitle

\begin{abstract}
We propose a method to improve subject transfer in motor imagery BCIs by aligning covariance matrices on a Riemannian manifold, followed by computing a new common spatial patterns (CSP) based spatial filter. We explore various ways to integrate information from multiple subjects and show improved performance compared to standard CSP. Across three datasets, our method shows marginal improvements over standard CSP; however, when training data are limited, the improvements become more significant.
\end{abstract}


\section{Introduction}
\label{sec:introduction}

Motor imagery BCIs translate user-generated control signals into mental states using pre-trained classifiers, which correspond with an action on some external interface, such as the movement of a mouse cursor. Common Spatial Patterns (CSP) is a commonly used supervised algorithm used to generate features by learning a projection matrix that maximizes the projected variance for one class while minimizing it for the other, enhancing discriminative power \cite{b5}. These classifiers are usually trained during a calibration session and applied to unseen data within or between sessions of the same subject.
EEG-based BCIs are especially challenging to implement because they are a non-stationary system \cite{b21}. Factors such as electrode impedance and decreasing user attention levels can degrade classifier performance over time \cite{b11}. Moreover, a motor imagery classifier trained on one subject's data cannot be reliably used on other subjects \cite{b12}.

\setlength{\parindent}{0pt}

To address the challenge of subject transfer, domain adaptation has been used to adapt classifiers for use across different subjects \cite{b2}. Approaches include adapting already extracted CSP features between subjects \cite{b6}; aligning covariance matrices on a Riemannian manifold and using a classifier on the aligned matrices directly \cite{b17}; and integrating subject transfer into CSP \cite{b7,b10,b18}. Throughout this work we refer to source and target subjects. In this context, the target refers to the subject for whom the model is being trained, while the source refers to other subjects whose data is leveraged to assist in training the model for the target subject.

\setlength{\parindent}{0pt}
In this work, we propose a novel method that combines the alignment of covariances on a Riemannian manifold with CSP, demonstrating its effectiveness across three datasets. We ultimately attain improved performance compared to Composite CSP \cite{b10}, a prominent method that uses transfer learning directly in CSP; Riemannian Procrustes Analysis \cite{b17}, a Riemannian Geometry based method for domain adaptation in BCI; and quite a pronounced performance increase over standard CSP in scenarios where limited training data is available.

\section{Related Work}
\label{sec:related-works}
\subsection{Riemannian Manifolds}
The space of \(n\times n\) symmetric positive definitive (SPD) matrices forms a Riemannian manifold, \(\mathcal{P}(n)\) \cite{b4}, which provides a more effective way to manipulate covariance matrices from subject motor imagery trials. We will define Riemannian manifolds via a chain of definitions and briefly discuss key properties relevant to this work.
\begin{itemize}
    \item \textbf{Topological Manifolds:} Spaces that look flat locally but are curved overall, like a circle or the Earth.

    \vspace{0.75em}
    
    \item \textbf{Differential Manifolds:} These are topological manifolds that are equipped with a differential structure, allowing us to map curved regions to flat spaces using bijective functions called charts, similar to how different parts of the curved Earth are mapped onto flat maps.

    \vspace{0.75em}
    
    \item \textbf{Smooth Manifolds:} These are differential manifolds where the transitions between charts are smooth, meaning no sharp changes. At any point \( X \) on \(\mathcal{P}(n)\), we can define a tangent space \( T_X\mathcal{P}(n) \), to perform linear approximations within the neighborhood of \(X\).

    \vspace{0.75em}
    
    \item \textbf{Riemannian Manifolds:} Smooth manifolds where each tangent space is equipped with a metric which provides a way to measure distances and angles on the manifold.
\end{itemize}
We can map a matrix \(X\) on the manifold to the tangent space at \(A\) using the logarithm map \(Log_A(.)\), and then map it back to the manifold using the exponential map \(Exp_A(.)\). Here, A denotes the matrix whose tangent space is considered.

\[
\tilde{X} = Log_A(X) = X^{1/2}log(X^{-1/2}AX^{-1/2})X^{1/2} 
\]
\vspace{0.1em}
\[
X = Exp_A(\tilde{X}) = \tilde{X}^{1/2}exp(\tilde{X}^{-1/2}A\tilde{X}^{-1/2})\tilde{X}^{1/2}
\]

Where \(log\) and \(exp\) refer to the matrix logarithm and exponential. By definition \(Log_A(A) = \boldsymbol{0}\), meaning that the tangent space of A is a coordinate system in which \(A\) is the origin. 

A common metric to equip every tangent space with is the following:

\[
\scalebox{0.9}{$\forall \tilde{X}^{(1)},\;\tilde{X}^{(2)}  \in T_A\mathcal{P}(n),\langle\tilde{X}^{(1)},\tilde{X}^{(2)}\rangle_A = \operatorname{tr}(A^{-1}\tilde{X}^{(1)}A^{-1}\tilde{X}^{(2)})$}
\]

Which is known as the Affine-Invariant Riemannian metric(AIRM) and \(tr(.)\) is the trace operator. From this metric we can define distances between two matrices on the manifold \(X^{(1)}\) and \(X^{(2)}\) as

\[
\delta_{AIRM}(X^{(1)},X^{(2)}) = \sum_{k=1}^n log^2(\lambda_k)
\]

Where \(\lambda_1,\dots ,\lambda_n\) are the eigenvalues of \(\left(X^{(1)}\right)^{-1}X_2\). With this definition of distance we can also determine the Riemannian Mean for a set of SPD matrices \(X^{(1)},...,X^{(n)}\) which is given as an optimization problem that can be solved by gradient descent \cite{b1}.
\[
\mu = \arg \min_{\mu \in \mathcal{M}} \sum_{i = 1}^{n} \delta_{\text{AIRM}}^2(X^{(i)}, \mu)
\]

\subsection{Common Spatial Patterns}
\label{sec:csp-explanation}
To obtain spatial filters with CSP, we first compute the average covariance matrix for each class \(c \in \{-1,+1\}\):
\[
\Sigma_{(c)} = \frac{1}{|I_c|}\sum_{i \in I_c}\frac{X^{(i)}X^{(i)'}}{tr(X^{(i)}X^{(i)'})}
\]
where \(I_c\) is the set of indices for class \(c\), and \(X^{(i)} \in \mathbb{R}^{C \times T}\) is a trial with \(C\) channels and \(T\) samples. The spatial filters \(W\) are the generalized eigenvectors from:
\[
W\Sigma^{(-1)} = \lambda(\Sigma_{(-1)}+\Sigma_{(+1)})W
\]
These filters are sorted by the generalized eigenvalues \(\lambda\) in descending order and \(n\) pairs are selected from the first and last \(n\) columns of \(W\). The feature vector extracted from a motor imagery trial \(X^{(i)}\) is:
\[
f_i = \log\left(diag\left\{W^{'}X^{(i)}X^{(i)'}W\right\}\right)
\]
\(diag(.)\) extracts diagonal elements as a vector, and \(log(.)\) is applied element-wise to this vector. In our implementation, we first apply a fifth-order Butterworth filter to bandpass the trials between 8-30 Hz, followed by the selection of three pairs of spatial filters.
\subsubsection{Multi-Class CSP}
In a multiclass setting, CSP is applied in a One-vs-Rest manner by training CSP filters for each class against the combined data of the other classes. The features extracted with these filters are used to train separate classifiers for each class. During prediction, each binary classifier outputs a probability, and the class with the highest associated probability is chosen as the predicted label \cite{b19}.

\subsection{Riemannian Geometry in BCIs}
\label{sec:rg-in-bcis}

Riemannian Geometry naturally aligns with BCI applications, as EEG from motor imagery trials can be effectively represented as covariance matrices. Unlike CSP-based classifiers, approaches such as \cite{b3} directly classify these trial covariances using a Riemannian Minimum Distance to Mean (RMDM) classifier. Some more recent approaches use neural networks and the Riemannian geometry of SPD matrices to better characterize the spectral and temporal features of EEG signals \cite{b9}. Beyond motor imagery classification, Riemannian geometry has also been used to analyze error related activity in BCIs \cite{b15}. It is still an open question whether or not these Riemannian approaches lead to improved performance; some work such as \cite{b8} suggest that the difference is minimal, however, there is a lack of a comprehensive investigation of old and new methods in the literature that adequately address this.

\subsection{Riemannian Geometry based Domain Adaptation}
\label{sec:rg-in-da-for-bci}

Earlier approaches to transfer learning and domain adaptation in motor imagery BCIs were simpler, with methods such as Composite CSP \cite{b10} that computed CSP-based spatial filters from a composite covariance matrix constructed from a linear combination of average class covariances from source and target subjects.
Recent advancements, such as Riemannian Procrustes Analysis (RPA) \cite{b17}, have integrated Riemannian geometry into domain adaptation for BCIs. RPA maps source and target covariances into a shared subspace and trains the previously mentioned RMDM classifier with these transformed source and target covariances. Some methods such as \cite{b20} further build on this approach and perform unsupervised alignment in a tangent space to create a common subspace of source and target data, and train a classifier on aligned and lower dimensional tangent space representations rather than on the trial covariances directly as in \cite{b17}.

\section{Riemannian Transfer CSP}
\label{sec:RTCSP}

We now present our method for subject transfer of CSP-based spatial filters which we refer to as Riemannian Transfer CSP (RTCSP). We provide a specific problem formulation, explain our method for a single subject transfer, then cover how this method can incorporate information from multiple subjects.

\subsection{Problem Setup}
\label{sec:problem-explanation}

We have a source subject S with N training motor imagery trials and their corresponding labels \(\{(X_S^{(i)},y_S^{(i)})\}_{i = 1}^N \) as well as a target subject T with M training trials and labels \(\{(X_T^{(i)},y_T^{(j)})\}_{j = 1}^M \). Each trial, \(X_S^{(i)}\) or \(X_T^{(j)}\) is a matrix in \(\mathbb{R}^{C\text{x}T}\), where C is the number of channels and T is the number of samples per channel. We will refer to the set of motor imagery trials associated with the source and target subject as \(\boldsymbol{X_S}\) and \(\boldsymbol{X_T}\) respectively. \(X_S^{(i)}\) will refer to the \(i^{th}\) motor imagery trial of source subject S and similarly \(X_T^{(j)}\) will refer to the \(j^{th}\) trial of target subject T. 

Associated with each trial is a covariance matrix in \(\mathbb{R}^{C\text{x}C}\). The covariance matrices of our source subject S are denoted as \(\boldsymbol{\Sigma_S}\) and as \(\boldsymbol{\Sigma_T}\) for our target subject. 
\[
\boldsymbol{\Sigma_S} = \left\{\frac{X_S^{(i)}X_S^{(i)'}}{tr\left( X_S^{(i)}X_S^{(i)'} \right)}\right\}_{i=1}^N 
\]

\[
\boldsymbol{\Sigma_T} = \left\{\frac{X_T^{(j)}X_T^{(j)'}}{tr\left( X_T^{(j)}X_T^{(j)'} \right)}\right\}_{j=1}^M
\]

Then our goal is to align the values of \(\boldsymbol{\Sigma_S}\) with \(\boldsymbol{\Sigma_T}\). After alignment we combine the aligned source covariances \(\boldsymbol{\Sigma_{S_{al}}}\) and the source labels with the target covariances and training labels. That is, \(\boldsymbol{\Sigma} =\boldsymbol{\Sigma_{S_{al}}}\) \(\cup\) \(\boldsymbol{\Sigma_T}\) and \(y = y_s\) \(\cup\) \(y_T\). We then run CSP on \(\boldsymbol{\Sigma}\) and \(y\) to obtain a spatial filter that we apply onto the train and test trials of the target subject.

\subsection{Single Subject Transfer}
\label{sec:single-subject-transfer}

The alignment process is explained for a single class and repeated for the other. First, we compute the Riemannian Means of \(\boldsymbol{\Sigma_S}\) and \(\boldsymbol{\Sigma_T}\), denoted \(M_S\) and \(M_T\). We then define the tangent spaces at these means, \(T_{M_S}\mathcal{P}(C)\) and \(T_{M_T}\mathcal{P}(C)\). Each covariance in \(\boldsymbol{\Sigma_S}\) and \(\boldsymbol{\Sigma_T}\) is mapped to its tangent space, where they are still \(C\) x \(C\) matrices. We then vectorize these matrices by flattening their upper-triangular elements. These \(C \times C\) matrices become vectors with \(\frac{C(C+1)}{2}\) elements.

\[
\tilde{\boldsymbol{\Sigma}}_{\boldsymbol{S}} = \left\{ \text{vec} \left(Log_{M_S}\left(\boldsymbol{\Sigma_S^{(i)}}\right) \right) \right\}_{i=1}^N
\]

\[
\tilde{\boldsymbol{\Sigma}}_{\boldsymbol{T}} = \left\{ \text{vec} \left(Log_{M_T}\left(\boldsymbol{\Sigma_T^{(j)}}\right) \right) \right\}_{j=1}^M
\]

vec is the operation that flattens the upper-triangular elements of a symmetric matrix into a vector.

Once we have \(\tilde{\boldsymbol{\Sigma}}_{\boldsymbol{S}}\) and \(\tilde{\boldsymbol{\Sigma}}_{\boldsymbol{T}}\), we obtain their top two principal components \(\boldsymbol{P_S}\) and \(\boldsymbol{P_T}\). These are the principal components of the covariance matrices mapped to their tangent spaces and flattened into vectors. We use the top two principal components for each. The idea is to align these vectors, turn them back into symmetric matrices, and then map them back to the manifold from the tangent space. To align these vectors, we begin by obtaining their Cholesky Decomposition's in their respective spaces spanned  by their top two principal components.

\[
    \boldsymbol{P_S}\tilde{\boldsymbol{\Sigma}}_{\boldsymbol{S}}\tilde{\boldsymbol{\Sigma}}_{\boldsymbol{S}}^{'}\boldsymbol{P_S^{'}} = \boldsymbol{L_SL_S^{'}}
\]

\[
    \boldsymbol{P_T}\tilde{\boldsymbol{\Sigma}}_{\boldsymbol{T}}\tilde{\boldsymbol{\Sigma}}_{\boldsymbol{T}}^{'}\boldsymbol{P_T}^{'} = \boldsymbol{L_T}\boldsymbol{L_T^{'}}
\]

\(\boldsymbol{P_S}\) and \(\boldsymbol{P_T}\) have shape 2 x \(\frac{C(C+1)}{2}\), thus the resulting  Cholesky factors \(\boldsymbol{L_S}\) and \(\boldsymbol{L_T}\) have shape 2 x 2. The Cholesky Decompositions are done in these two dimensional principal component spaces because when we go from the \(C\)x\(C\) covariance matrices flattened into vectors with \(\frac{C(C+1)}{2}\) elements, often \(\frac{C(C+1)}{2}\) is much greater than the number of samples available, which can result in ill-conditioned covariance matrices and consequently unstable Cholesky Decompositions. We are then able to align the values of \(\tilde{\boldsymbol{\Sigma}}_{\boldsymbol{S}}\) with \(\tilde{\boldsymbol{\Sigma}}_{\boldsymbol{T}}\) with the transformation
\[
\tilde{\boldsymbol{\Sigma}}^{'}_{\boldsymbol{S_{al}}} = \left\{\boldsymbol{P_T^{'}L_TL_S^{-1}P_S}\tilde{\boldsymbol{\Sigma}}_{\boldsymbol{S}}^{(i)'}\right\}_{i=1}^N
\]

Where \(\tilde{\boldsymbol{\Sigma}}_S\) has shape \(N\times\frac{C(C+1)}{2}\) and N is the number of training samples available for a class of source subject S. Then we can map the elements of \(\tilde{\boldsymbol{\Sigma}}_{\boldsymbol{S_{al}}}\) back onto the manifold by turning them back into symmetric matrices and using the exponential map defined about \(M_T\).

\[
\boldsymbol{\Sigma_{S_{al}}} = \left\{ Exp_{M_T}\left( \text{mat}\left( \boldsymbol{\tilde{\Sigma}_{S_{al}}^{(i)}} \right) \right) \right\}_{i=1}^N
\]

Here, mat transforms a vector of \(\frac{C(C+1)}{2}\) elements into a symmetric \(C \times C\) matrix. After repeating this for both classes, we compute a new spatial filter by applying CSP on \(\boldsymbol{\Sigma} = \boldsymbol{\Sigma_{S_{al}}} \cup \boldsymbol{\Sigma_T}\) and \(y = y_s \cup y_T\), resulting in the spatial filter \(W_S\), as in the spatial filter transferred from source subject \(S\). \(W_S\) is applied to the target subject's training trials, yielding the log-variance features \(f_S\).

\[
f_S = \left\{\log\left(\text{diag}\left(W_S^{'}X_T^{(j)}X_T^{(j)}{'}W_S\right)\right)\right\}_{j=1}^M
\]

The features \(f_S\) are paired with the target subject training labels \(y_T\) and are used to train an LDA classifier. The same spatial filter \(W_S\) is used to get log variance features from the target subject test data, and the offline-trained LDA classifier can then make predictions on these features. We emphasize that this is the process for a single subject transfer, and we expand on this so that we can incorporate information from multiple source subjects. For the multi-class case this alignment process is simply repeated in a One-versus-Rest manner.

\subsection{Multiple Subject Transfer}
\label{sec:multiple-subject-transfer}

We suggest three different ways to incorporate information from different subjects into our transfer paradigm that involve generating a single spatial filter from multiple subjects, generating multiple sets of features, and an ensemble approach.

\subsubsection{RTCSP Single Spatial Filter}
\label{sec:RTCSP-SSF}

We first present a method that computes a single spatial filter using the information from all source subjects denoted as RTCSP Single-Spatial-Filter (RTCSP-SSF). Consider we have a total of K source subjects \(S_1,\dots\ ,S_K\) and their associated set of covariances \(\boldsymbol{\Sigma_{S_1}},\dots\ ,\boldsymbol{\Sigma_{S_K}}\) as well as a single target subject \(S_T\) and a set of target covariances \(\boldsymbol{\Sigma_T}\). Using our alignment process, we align each set of source covariances with \(\boldsymbol{\Sigma_T}\) and get \(\boldsymbol{\Sigma_{S_1^{al}}},\dots\ ,\boldsymbol{\Sigma_{S_K^{al}}}\). We combine all of the covariances from the target subject and the aligned covariances from the source subjects as well as their labels in the following manner,

\[
    \boldsymbol{\Sigma} = \boldsymbol{\Sigma_T} \text{ } \cup \text{ } \boldsymbol{\Sigma_{S_1^{al}}} \text{ } \cup \dots\ \cup \boldsymbol{\Sigma_{S_K^{al}}}
\]
\[
    y = y_{T} \text{ } \cup y_{S_1} \text{ } \cup \dots\ \cup \text{ } y_{S_K}  
\]

We run CSP on \(\boldsymbol{\Sigma}\) and \(y\) to get a spatial filter. In the previously described manner, we can use this spatial filter to get log-variance features in the CSP space from the training trials of the target subject, and pair these features with the target subject training labels \(y_T\) to train an LDA classifier. This spatial filter is also then applied on the unseen test trials of the target subject to get log-variance features during test time and the LDA classifier that was trained offline can then make predictions on these feature vectors.

\subsubsection{RTCSP-Combine}
\label{sec:RTCSP-Combine}

We use the same setup with \(K\) source subjects \(S_1, \dots, S_K\) and their covariances \(\boldsymbol{\Sigma_{S_1}}, \dots, \boldsymbol{\Sigma_{S_K}}\), along with target subject \(S_T\) and its covariances \(\boldsymbol{\Sigma_T}\). After alignment, we obtain \(\boldsymbol{\Sigma_{S_1^{al}}}, \dots, \boldsymbol{\Sigma_{S_K^{al}}}\), and instead of merging these, as in the RTCSP-SSF approach, we use them separately to get individual spatial filters \(W_T,W_{S_1}, \dots, W_{S_K}\) to extract multiple sets of log-variance features.

\begin{algorithm}[H]
  \caption{RTCSP-Combine}
  \textbf{Input}: Source subjects \(S_1,\dots, S_K\)\\
  \hspace*{2.75em} Target subject \(S_T\)
  \begin{algorithmic}
  
    \STATE Initialize \texttt{TrainFeatures = []}
    \STATE Initialize \texttt{TrainLabels = []}
    
    \FOR{$i = 1$ \textbf{to} $K$}
      \STATE $\Sigma_{S_i^{\text{al}}} = \texttt{Align}(\Sigma_{S_i},\Sigma_T)$
      \STATE $W_{S_i} = \texttt{CSP}(\Sigma_{S_i^{\text{al}}},y_{S_i})$
      \STATE $f_{S_i} = \texttt{getLogVarFeatures}(X_T,W_{S_i})$
      \STATE \texttt{TrainFeatures.append}($f_{S_i}$)
      \STATE \texttt{TrainLabels.append}($y_T$)
    \ENDFOR
    
    \STATE $W_T = \texttt{CSP}(\Sigma_T,y_T)$
    \STATE $f_T = \texttt{getLogVarFeatures}(X_T,W_T)$
    
    \STATE \texttt{TrainFeatures.append}($f_T$)
    \STATE \texttt{TrainLabels.append}($y_T$)
    \STATE \texttt{clf} = \texttt{LDA(TrainFeatures, TrainLabels)}
  \end{algorithmic}
\end{algorithm}

The key difference here from RTCSP-SSF is that we are learning multiple CSP based spatial filters \(W_{S_1}\text{, } \dots\ \text{,}W_{S_k}\) that have been aligned with the spatial filter of our target subject \(W_T\). We emphasize that these spatial filters are being used to generate new data points and \textit{not} being used to generate extra dimensions for the LDA classifier. This is done by applying each CSP based spatial filter onto \(X_T\). If our target subject has some original \(M\) training examples, we use \(W_{S_1}\) on \(X_T\) and generate an additional M training examples, then we apply \(W_{S_2}\), and generate another M training examples. This is done for all of the spatial filters \(W_{S_1},\dots\,W_{S_K}\), giving a total of  \(M(K+1)\) training examples  to train the LDA classifier. If the original motor imagery trials have C channels and we pick n pairs of eigenvectors for each spatial filter, then each \(W_{S_i}\) will be in \(\mathbb{R}^{C\text{x}2n}\). When we use each \(W_{S_i}\) on the training trials of the target subject, \(X_T\), then the resulting features will be in \(\mathbb{R}^{2n}\). We argue that the resulting features can simply be combined like this because they are all drawn from the same distribution since before the spatial filters are used to generate these features, they are all aligned with the same target subject.
 
For making predictions on unseen test trials, it does not make practical sense to increase the size of the test dataset as we did during training, thus in test time we use the spatial filter from the first explained method, RTCSP-SSF, on test data.

\subsubsection{RTCSP-Ensemble}
\label{sec:RTCSP-ENS}

We also suggest an ensembling approach using the multiple sets of generated log-variance features but in a rather specific manner.

Again we obtain a set of spatial filters \(W_{S_1}, \dots, W_{S_K}\), each corresponding to one of the source subjects. These spatial filters are  applied to the training trials of the target subject, generating multiple sets of log-variance features denoted as \(f_{S_1}, \dots, f_{S_K}\), where \(f_{S_i}\) represents the log-variance features derived using the spatial filter \(W_{S_i}\). 

To train our ensemble the \(i^{th}\) classifier in the ensemble is trained on the corresponding log-variance features \(f_{S_i}\) and the target subject training labels \(y_T\). We illustrate this process below in a more concrete manner in terms of an algorithm.

\begin{algorithm}[H]
  \caption{RTCSP-Ensemble}
  \textbf{Input}: Source subjects \(S_1, \dots, S_K\) \\
  \hspace*{2.75em} Target subject \(S_T\)

  \begin{algorithmic}
    \STATE Initialize \texttt{EnsembleClf = []}
  
    \FOR{$i = 1$ \textbf{to} $K$}
      \STATE \(\Sigma_{S_i^{al}} = \texttt{Align}(\Sigma_{S_i}, \Sigma_T)\)
      \STATE \(W_{S_i} = \texttt{CSP}(\Sigma_{S_i^{al}}, y_{S_i})\)
      \STATE \(f_{S_i} = \texttt{getLogVarFeatures}(X_T, W_{S_i})\)
      \STATE \texttt{clf} = \texttt{LDA}(\(f_{S_i},y_T\))
      \STATE \texttt{EnsembleClf.append(clf)}
    \ENDFOR
    \STATE \(W_T = \texttt{CSP}(\Sigma_T,y_T)\)
    \STATE \(f_T = \texttt{getLogVarFeatures}(X_T,W_T)\)
    \STATE \texttt{TargetClf = LDA}(\(f_T,y_T\))
    \STATE \texttt{EnsembleClf.append(TargetClf)}
  \end{algorithmic}  
\end{algorithm}

For an unseen motor imagery test trial \(X\), spatial filters \(W_T, W_{S_1}, \dots, W_{S_K}\) extract log-variance features using \(\log(\text{diag}(W_{S_i}^{'} X X^{'} W_{S_i}))\). Each feature is classified by the corresponding classifier, and the final label is determined by a majority vote.

\section{Results}
\label{sec:Results}

We present our results across three different motor imagery datasets and also with a custom experiment showing how our method compares to the baseline CSP method when different amounts of training data are available for the target subject. 

\subsection{Datasets}
\label{sec:datasets-desc}

\begin{enumerate}
    \item \textbf{BCI Competition III Dataset IVa:} A two-class(left hand, right foot) motor imagery dataset with five subjects (aa, al, aw, av, ay) having 168, 224, 56, 84, and 28 trials respectively. The channel count was  reduced from 118 to 68 to avoid rank deficiency in covariance matrices as done in\cite{b18}. Additionally, Ledoit-wolf shrinkage is used on the resulting covariances.
    
    \item \textbf{BCI Competition IV Dataset 2a:} A four-class(left hand, right hand, tongue, and foot) motor imagery dataset recorded in \cite{b16} with 22 channels and 72 trials per class per subject, excluding artifactual trials.

    \item \textbf{Dataset Three:} A two-class(left hand,right hand) motor imagery dataset used in \cite{b13} and \cite{b14} with 64 channels and 136 trials per subject for 12 subjects. Ledoit-wolf shrinkage is also used here on trial covariances.
    
\end{enumerate}

\subsection{Compared Methods}
\label{sec:compared-methods}

\begin{enumerate}

\item \textbf{Composite CSP} (cCSP): Computes a spatial filter using a linear combination of average class covariances from the target and source subjects, which is then used to obtain a spatial filter that is applied to both training and test trials of a target subject.

\item \textbf{Riemannian Procrustes Analysis} (RPA): Aligns source covariances to the target subject’s covariances via Riemannian Geometry. Rather than creating a new CSP-based filter, aligned covariances are used with a Riemannian Minimum Distance to Mean (RMDM) classifier, trained on both aligned source and target covariances.

\end{enumerate}

\subsection{Hyper-Parameter Tuning}
\label{sec:ht-tuning}

Hyper-parameter tuning was performed only for the Composite CSP method on each of the three benchmark datasets. Tuning was done exclusively with training data, using train-validation splits for the target subject's training data, while all source subject data was used. The optimal \(\lambda\), minimizing average validation error across splits, was used to obtain the Composite CSP with all source and target subject training data.

\begin{enumerate}
    \item \textbf{BCIC IV Dataset 2A and Dataset Three} To tune the \(\lambda\) parameter in Composite CSP, we performed 10-Fold Cross Validation with \(\lambda\) values ranging from 0.1 to 0.9. 

    \item \textbf{BCIC III Dataset 4a:} Given the varying number of training trials in this dataset, \(\lambda\) was tuned using Leave-One-Out Cross Validation. 

\end{enumerate}

\begin{figure}[H]
    \centering
    \includegraphics[width = 0.6\textwidth]{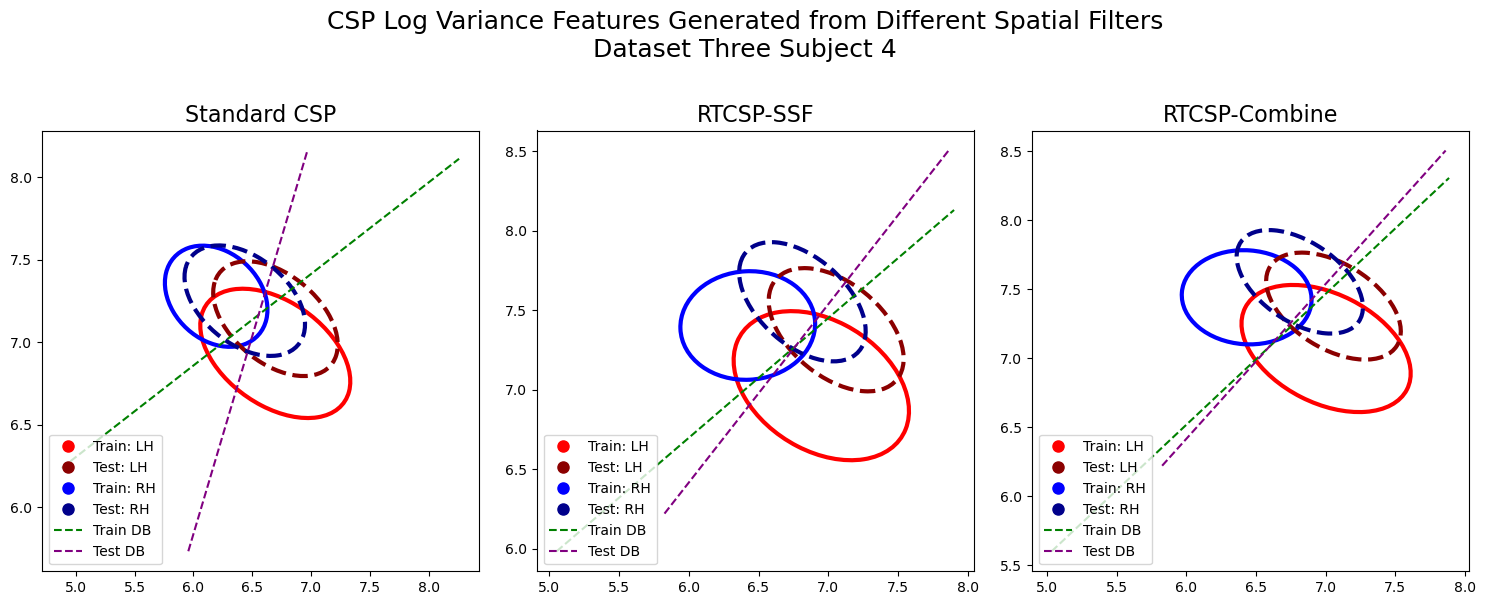}

    \includegraphics[width = 0.6\textwidth]{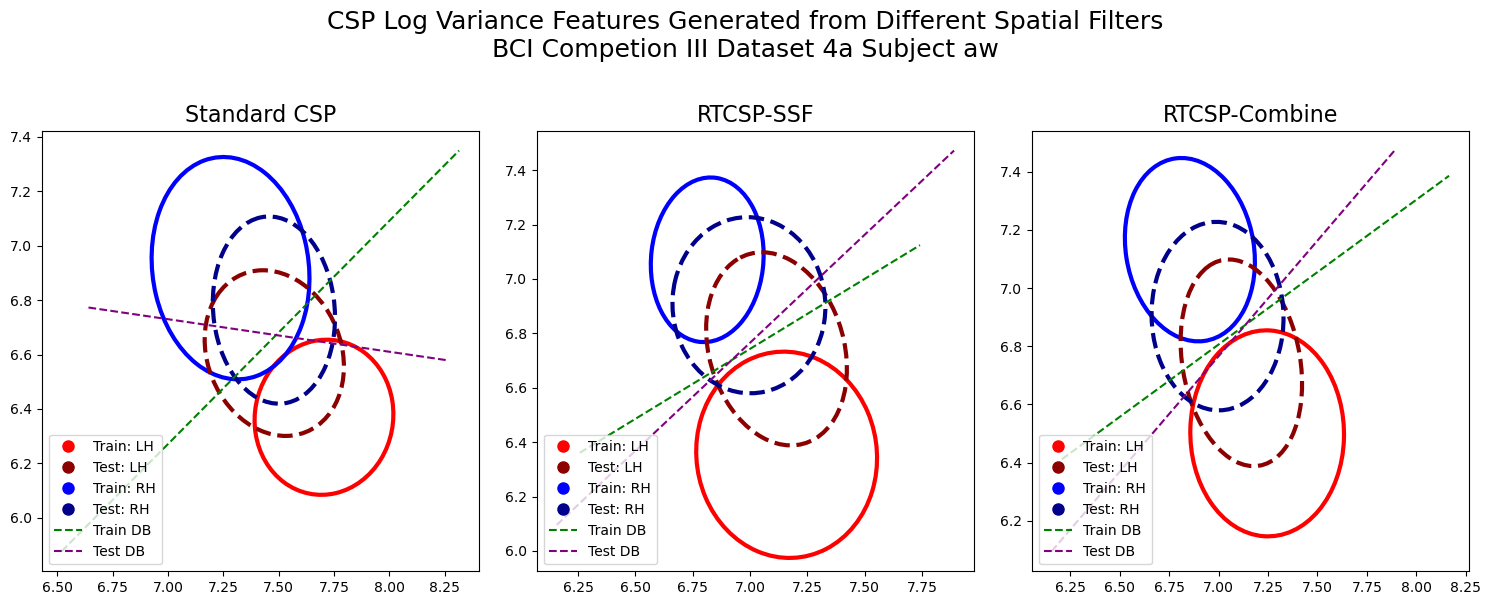}

    \onecolumn
    \caption{{{A comparison of features resulting from standard CSP with RTCSP-SSF and RTCSP-Combine. The spatial filter in each plot was obtained with training data and the same filter was used on both train and test data to obtain the resulting features. The ellipses represent the contours of the bi-variate normal distribution of features associated with each class of motor imagery at one standard deviation per axis. Ellipses associated with training data are solid colors, while ellipses associated with test data are the same color, but darker and with dashed lines. Decision boundaries for both the training and test data are shown. Test labels were only accessed for the purposes of this figure.}}}

    \label{sec:data-ellipses-figure}

\end{figure}

\subsection{Spatial Pattern Analysis}
\begin{figure}[H]
\centering
\includegraphics[width=1\textwidth]{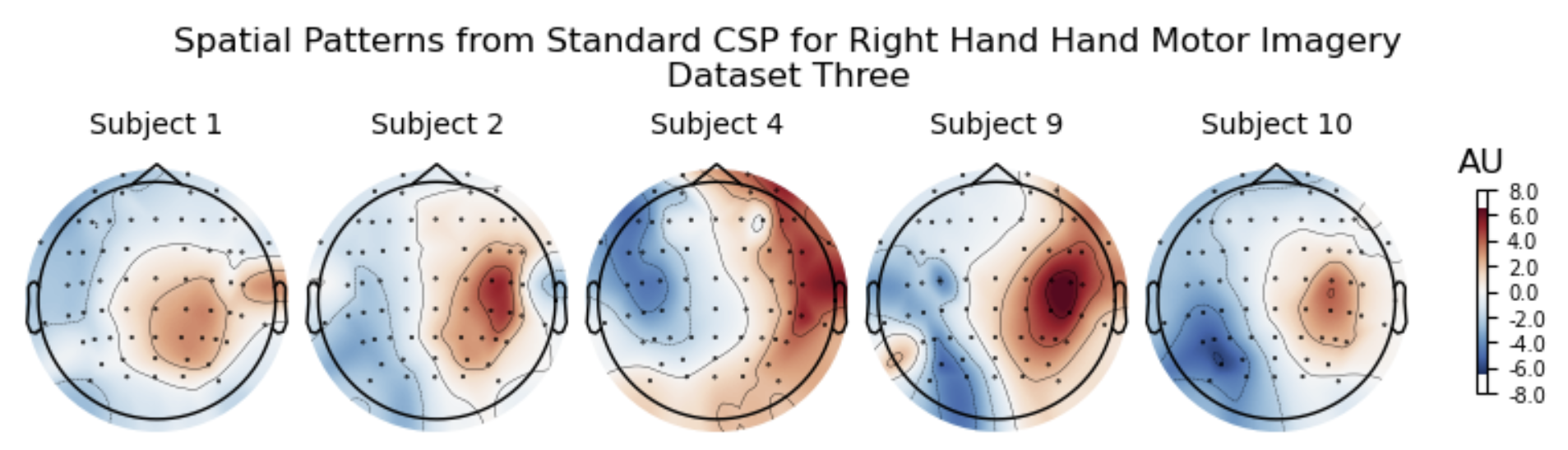}

\includegraphics[width=1\textwidth]{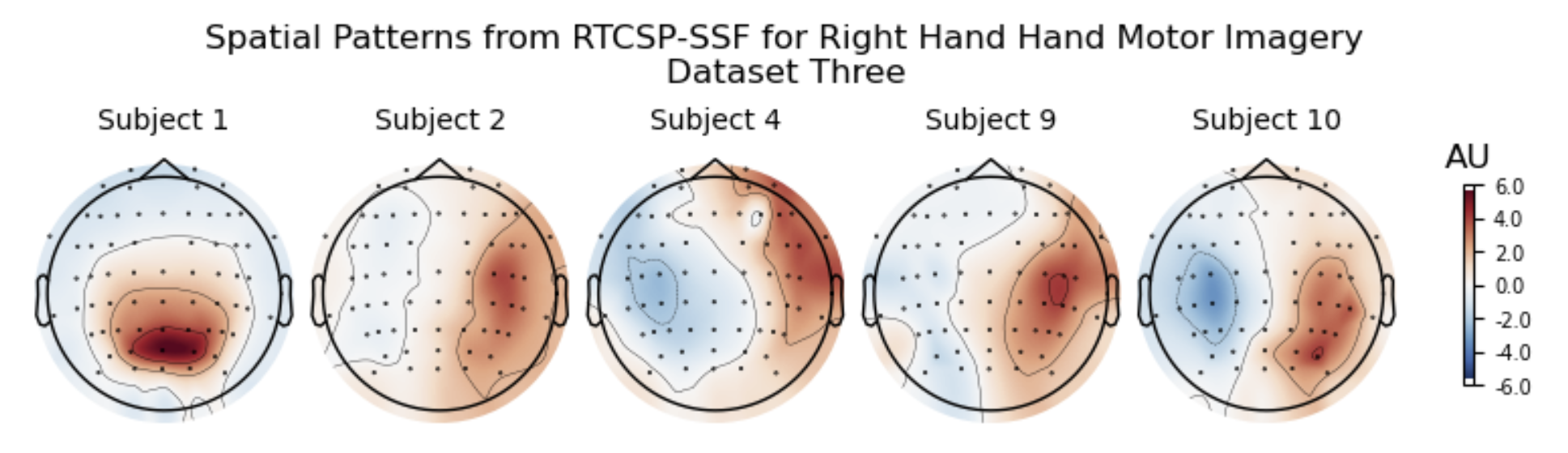}

\includegraphics[width=1\textwidth]{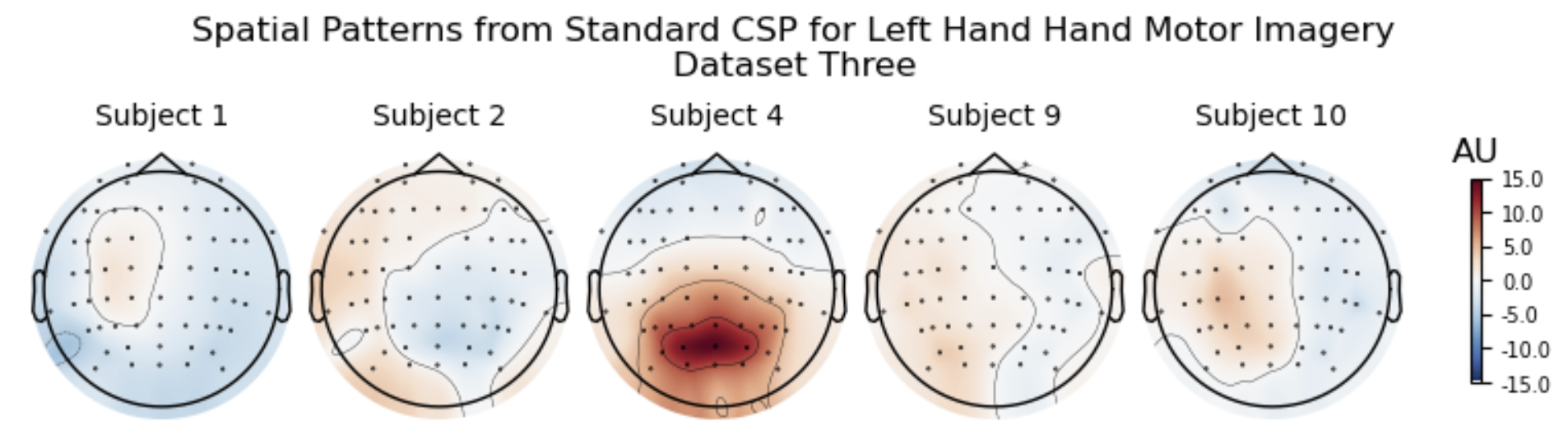}

\includegraphics[width=1\textwidth]{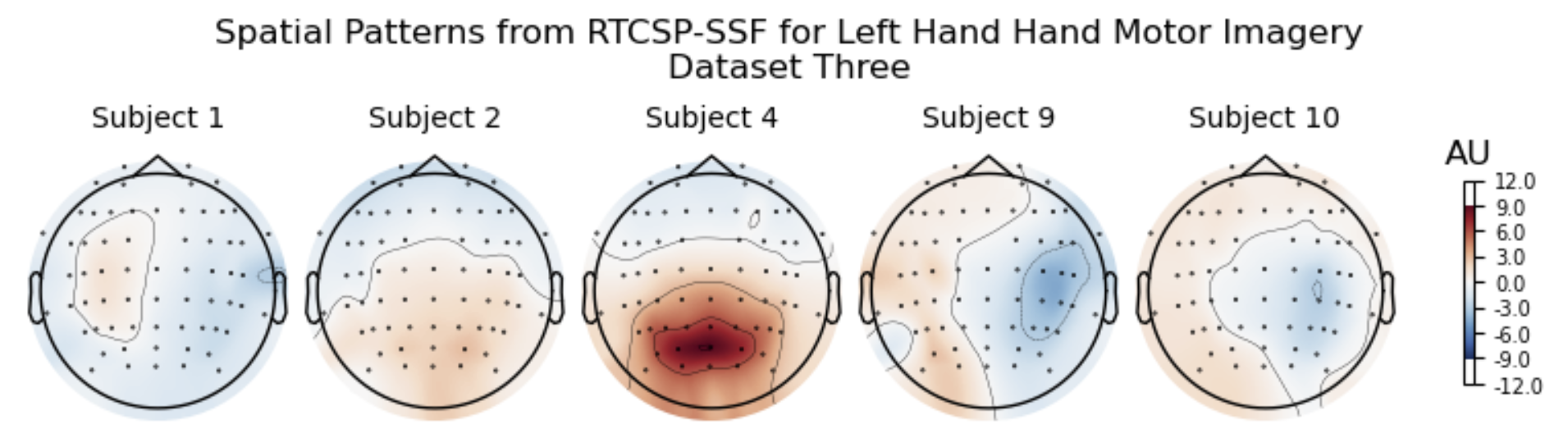}

\caption{Spatial patterns for standard CSP and RTCSP-SSF, showing how presumed neural sources project onto the scalp at each electrode. Polarity reflects the relative contribution of each source to the measured signals in sensor space. Generalized eigenvalues from the CSP problem were ordered from highest to lowest, and we plot one spatial filter associated with the three largest eigenvalues, and one filter from the three smallest.}
\end{figure}

\setlength{\arrayrulewidth}{1.5pt}
\begin{table}
    \centering
    \footnotesize
    \begin{tabular}{lccc|ccc} 
        \toprule
        \multicolumn{7}{c}{\textbf{BCIC IV Dataset 2a Test Accuracies (\%)}} \\ 
        \midrule
        \textbf{Subject} & \textbf{\makebox[2.3cm]{\shortstack{RTCSP\\SSF}}} & \textbf{\makebox[2.3cm]{\shortstack{RTCSP\\COMB}}} & \textbf{\makebox[2.3cm]{\shortstack{RTCSP\\ENS}}} & \textbf{CSP} & \textbf{cCSP} & \textbf{RPA} \\
        \midrule
        1 & 74.7 & 74.7 & 74.3 & 72.9 & 74.0  & \textbf{77.4} \\
        2 & \textbf{45.5} & \textbf{45.5} & \textbf{45.5} & 44.1 & 44.4 & 32.6 \\
        3 & 79.5 & 78.8 & 79.2 & \textbf{81.6} & 80.6 & 70.8 \\
        4 & 62.2 & 63.5 & 62.5 & \textbf{64.2} & 63.5 & 47.2 \\
        5 & 38.5 & 39.2 & 38.5 & 39.9 & \textbf{41.0} & 36.5 \\
        6 & 46.5 & \textbf{46.9} & 46.2 & 43.1 & 42.4 & 40.3 \\
        7 & 78.5 & \textbf{79.2} & 78.5 & 73.6 & 75.0 & 67.4 \\
        8 & 76.7 & 77.1 & 77.4 & 78.5 & 78.1 & \textbf{77.8} \\
        9 & 72.6 & 73.3 & \textbf{74.0} & 70.8 & 73.2 & 72.6 \\
        \midrule
        \rule{0pt}{3ex} 
        Mean & 63.9 & \textbf{64.2} & 64.0 & 63.2 & 63.6 & 58.0 \\
        \bottomrule
    \end{tabular}
    \vspace{1em}
    \caption{Performance of Different methods on BCI Competition IV Dataset 2a. The best result for each subject is shown in bold. The columns to the left of the dashed line include our proposed three methods. Test accuracy refers to methods performance on unseen data. This dataset had four classes. Our methods tend to perform marginally better than standard CSP.}
    
    \label{sec:BCIC-IV-D2A-full-data-results}
\end{table}

\begin{table}[H]
    \centering
    \footnotesize
    \begin{tabular}{lccc|ccc} 
        \toprule
        \multicolumn{7}{c}{\textbf{BCIC III Dataset 4a Test Accuracies(\%)}} \\ 
        \midrule
        \textbf{Subject} & \textbf{\makebox[2.3cm]{\shortstack{RTCSP\\SSF}}} & \textbf{\makebox[2.3cm]{\shortstack{RTCSP\\COMB}}} & \textbf{\makebox[2.3cm]{\shortstack{RTCSP\\ENS}}} & \textbf{CSP} & \textbf{cCSP} & \textbf{RPA} \\
        \midrule
        aa &\textbf{69.6} & 67.8 & 68.8 & 62.5 & 64.3 & 57.1 \\
        aw & 65.2 & 62.5 & \textbf{66.1} & 60.7 & 60.2 & 52.2 \\
        av & 59.2 & 57.1 & 59.7 & 59.2 & \textbf{60.2} & 59.7 \\
        al & 71.4 & 71.4 & 71.4 & 76.8 & 78.5 & \textbf{89.3} \\
        ay & 73.4 & 73.8 & \textbf{74.2} & 71.0 & \textbf{74.2} & 50.4 \\
        \midrule
        \rule{0pt}{3ex}
        Mean & 67.8 & 66.5 & \textbf{68.0} & 66.0 & 67.5 & 61.8 \\
        \bottomrule
    \end{tabular}
    \vspace{1em}
    \caption{Performance of Different methods on BCI Competition III Dataset 4a. The best result for each subject is shown in bold. }
    \label{sec:BCIC-III-D4A-full-data-results}
\end{table}

\begin{table}[H]
    \centering
    \footnotesize 
    \begin{tabular}{lccc|ccc} 
        \toprule
        \multicolumn{7}{c}{\textbf{Dataset Three Test Accuracies (\%)}} \\ 
        \midrule
        \textbf{Subject} & \textbf{\makebox[2.3cm]{\shortstack{RTCSP\\SSF}}} & \textbf{\makebox[2.3cm]{\shortstack{RTCSP\\COMB}}} & \textbf{\makebox[2.3cm]{\shortstack{RTCSP\\ENS}}} & \textbf{CSP} & \textbf{cCSP} & \textbf{RPA} \\
        \midrule
        1 & 55.9 & 55.4 & 57.8 & 58.2 & 54.1 & \textbf{61.3} \\
        2 & \textbf{52.6} & 52.3 & \textbf{52.6} & 48.5 & 51.5 & 47.6 \\
        3 & 57.9 & 51.3 & 57.9 & \textbf{61.8} & 54.0 & 48.6 \\
        4 & 65.4 & 64.2 & 64.6 & 60.8 & \textbf{66.2} & 63.6 \\
        5 & 81.4 & 81.4 & \textbf{82.1} & 77.9 & 69.3 & 63.6 \\
        6 & \textbf{70.8} & 70.6 & 70.2 & 68.3 & 56.2 & 60.9 \\
        7 & 57.3 & 57.6 & 56.3 & 57.6 & 51.7 & \textbf{58.7} \\
        8 & \textbf{62.9} & 62.5 & 62.5 & 60.5 & 56.0 & 54.0 \\
        9 & 69.0 & 68.5 & 69.0 & 65.5 & \textbf{70.5} & 61.0 \\
        10 & 58.5 & 57.9 & 57.7 & 60.2 & 57.9 & \textbf{63.8} \\
        11 & 46.4 & \textbf{48.6} & 46.4 & 42.9 & 47.7 & 50.0 \\
        12 & 57.7 & 57.4 & 57.4 & \textbf{59.8} & 50.3 & 55.6 \\
        \midrule
        \rule{0pt}{3ex} 
        Mean & \textbf{61.3} & 60.7 & 61.2 & 60.1 & 57.1 & 57.2 \\
        \bottomrule
    \end{tabular}
    \vspace{1em}
    \caption{{Performance of Different methods on Dataset Three. Within this dataset, training trials were obtained with user feedback, while unseen test trials were obtained without feedback.}}

    \label{sec:dataset-three-full-data-results}
\end{table}

Overall, our method demonstrates slightly better performance than standard CSP across three benchmark datasets. We can see based on figure 1 that our methods compared to standard CSP tend to generate a decision boundary that is closer to the optimal one for the space of features that are extracted from unseen test data. Additionally, figure two allows us to verify the neurophysiological plausibility of RTCSP-SSF, which also extends to RTCSP-Combine and RTCSP-Ensemble. The resulting spatial patterns are characteristic of left and right-hand motor imagery and resemble those generated by the standard CSP method.

Additionally, we outperform both Composite CSP (cCSP) and Riemannian Procrustes Analysis (RPA). The improvement over RPA is particularly intriguing, given the methodological similarities. Both approaches involve basic linear transformations of source and target covariances; however, the transfer process in RPA differs in three key aspects.

The first difference lies in the complexity of RPA, which operates directly on the manifold (instead of the tangent space) using a gradient descent procedure to determine the optimal rotation for aligning source and target covariances. 

The second difference is that RPA aligns covariances on a high-dimensional manifold, while our method performs alignment on 
a two-dimensional approximation of the tangent space.

Lastly, RPA transforms both the source and target covariances, whereas our method leaves the target data unchanged and focuses solely on aligning the source data to the target data. 

A theoretical concern is that aligning in a two-dimensional approximation to the tangent space depends on the quality of the linear approximations, which, if poor, may not ensure proper alignment on the manifold. Empirically, however, tangent space alignment performs better and is simpler.

\subsection{Limiting Target Training Data}
\label{sec:limited-training-data-results}

Of interest is how much training data a classifier needs to generalize on unseen motor imagery trials and how quickly it can be calibrated. To address this, we compare our methods to standard CSP with varying amounts of target subject training data.

\begin{figure}[H]
    \centering

    \begin{subfigure}[t]{0.32\textwidth}
        \centering
        \includegraphics[height=5cm]{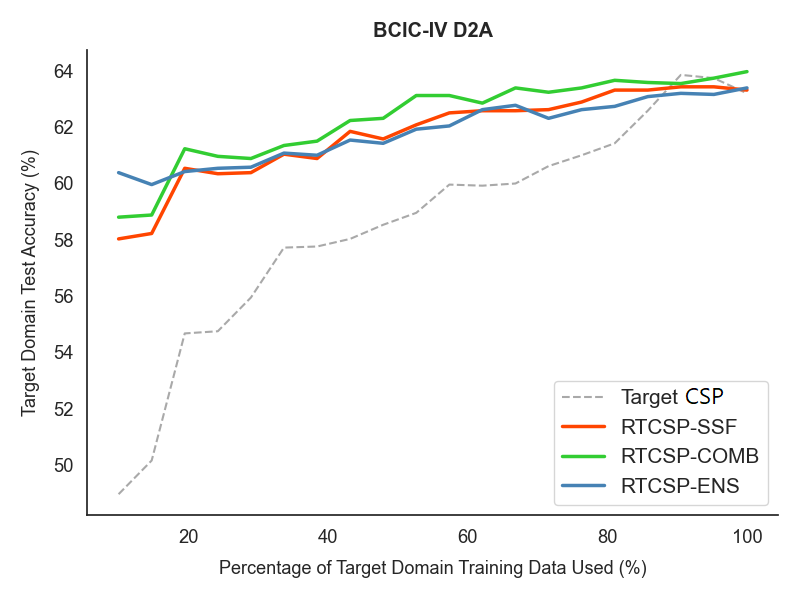}
    \end{subfigure}
    \hfill
    \begin{subfigure}[t]{0.32\textwidth}
        \centering
        \includegraphics[height=5cm]{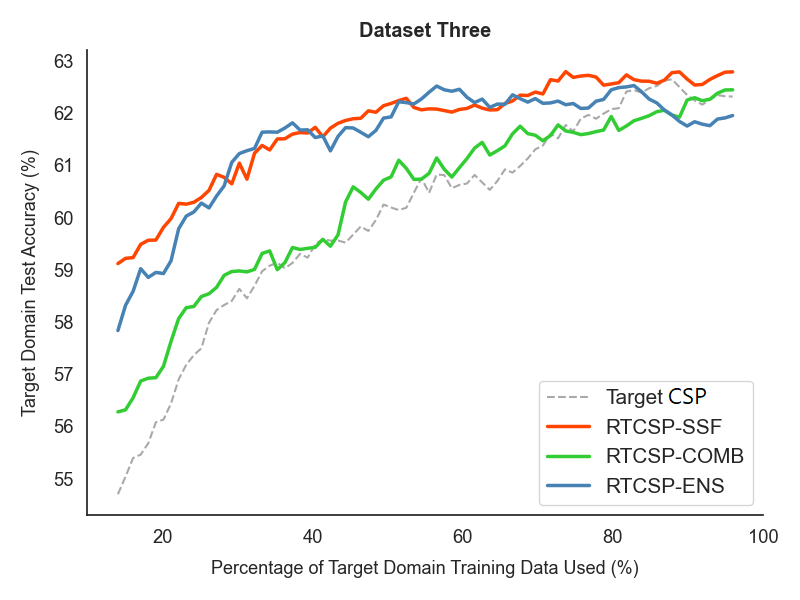}
    \end{subfigure}
    \hfill
    \begin{subfigure}[t]{0.32\textwidth}
        \centering
        \includegraphics[height=5cm]{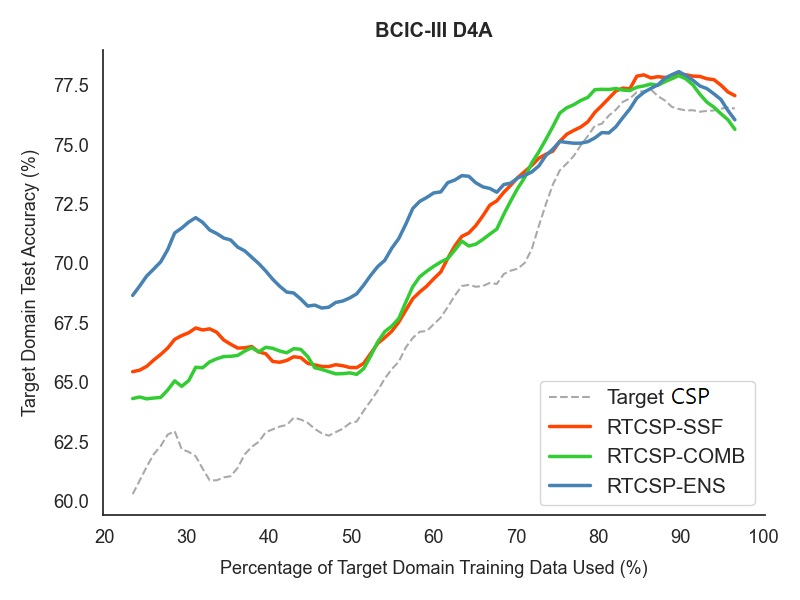}
    \end{subfigure}

    \caption{{Performance of our methods when different amounts of training data are available for the target subject. Our methods generalize to unseen data in low training data regimes more effectively than standard CSP.}}
    \label{sec:curves-less-training-data}
\end{figure}

Figure 3 shows improved performance over standard CSP with limited training data, indicating better generalization to unseen motor imagery trials. However, no consistently optimal method emerges across these datasets in low-data scenarios.

When generating the above curves for BCIC-IV D2A, we began with 10\%\ of training data, which was usually around 20-28 trials. In each iteration of this experiment this was increased by 5\%. The same type of curve was generated for each subject, and the displayed one is the average of these curves across subjects. 

For Dataset Three we started with 10\%\ of training data, which was 10 trials, and performed this experiment in 1\%\ increments. The original curve was rather noisy so in order to better visualize the long term trends, we smoothed out the curve with a moving average filter with a window size of 9.

For dataset BCIC-III D4A this curve started at 20\%\ and went up in 1\%\ increments. This curve was also noisy and we again used a moving average filter with a window size of 8. The rule of thumb we used for the window size was to divide the number of samples we have to plot by 10. 

\begin{figure}[H]
\centering

\includegraphics[width=0.7\textwidth]{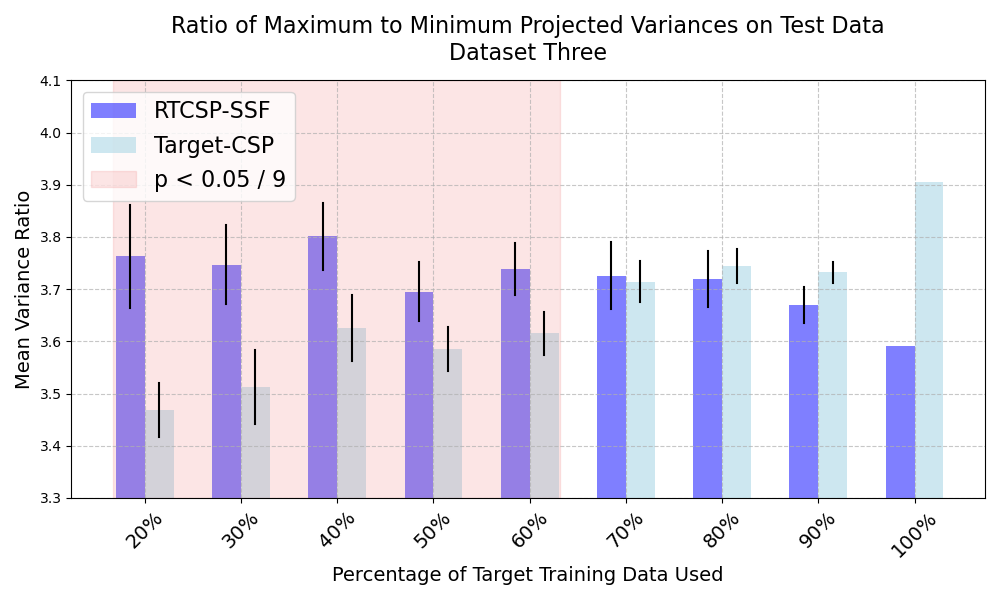}

\includegraphics[width=0.7\textwidth]{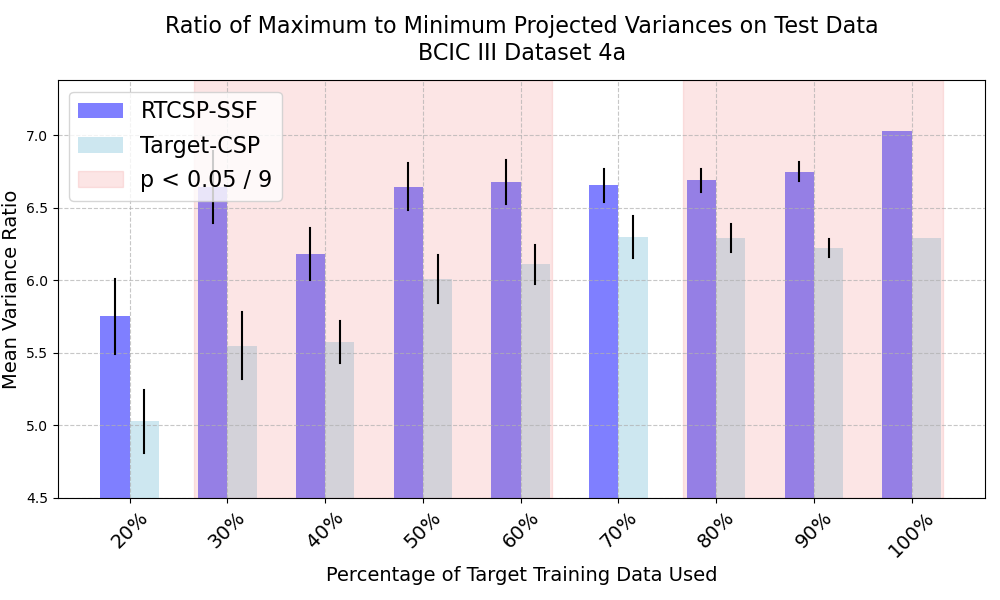}

\caption{{The ratio of maximum to minimum projected variances on datasets three and BCIC-III 4a, averaged over trials and subjects. This quantity reflects a spatial filter's ability to generalize and create discriminative features on unseen data, enabling more accurate classification. Standard CSP requires a larger amount of training data to generalize effectively, whereas RTCSP-SSF achieves similar performance with substantially less data. Error bars represent the standard error across 50 runs. Refer to Fig. 5 for the distribution corresponding to the left-most bars (20\% of training data used).}}
\end{figure}

As shown in Figure 3, classifiers trained with standard CSP features perform worse than those trained with features from RTCSP-SSF, especially when the amount of training data is limited. We attribute this to standard CSP's tendency to overfit, generating spatial filters that capture discriminative features from the training data but fail to generalize to unseen motor imagery trials. This overfitting results in variance-based features that are poorly separable during testing. In contrast, we hypothesize that RTCSP-SSF leverages aligned covariances from other subjects to produce spatial filters that generalize more effectively to unseen data. This hypothesis is further supported by the results in Figure 4, which demonstrate RTCSP-SSF's ability to create more discriminative features than standard CSP across varying amounts of training data. Appendix A should be referred to for the full details of this experiment.

From the results of the experiment we notice that the standard CSP algorithm is able to gradually create more discriminative features on test data the more training data is used, while RTCSP-SSF has the same capacity to generate these discriminative features even with considerably less training data. This difference between standard CSP and RTCSP-SSF is the most pronounced in more extreme conditions, such as when 20\% or 30\% of the target subject's training data is available. As more training data is available, the difference between RTCSP-SSF and standard CSP begins to diminish, which helps to explain the marginal improvements on our experiments with benchmark datasets.

This reasoning for improved performance extends naturally to RTCSP-Combine as well, as it employs the same spatial filter derived from RTCSP-SSF during testing on unseen data. The key difference lies in how the training data is obtained. Instead of pooling all covariances to produce a single spatial filter as done in RTCSP-SSF, RTCSP-Combine generates multiple spatial filters, with one for each source subject. This approach increases the amount of training data available for the LDA classifier, enhancing its ability to generalize while still leveraging the RTCSP-SSF filter for unseen data. Similarly, for RTCSP-Ensemble, it is unsurprising that an ensemble of weak classifiers leverages multiple sets of features to outperform standard CSP, particularly in low-data scenarios.

\section{Discussion}

In general our method shows marginal improvements in a scenario where we have a full set of training data that can be used to train an offline classifier. Where our method does particularly well is in scenarios where the target subject has limited training data. This suggests that this could be a useful method for faster calibration for online training of motor imagery classifiers should there be offline recorded and labeled data of the same format that this alignment process could use. 

While this method offers improvements over traditional CSP, especially when training data is limited, its efficiency in online settings remains suboptimal. The computation of Riemannian means, if performed continuously in real-time, may be impractical, since it is a gradient-descent based procedure with \textit{O}\((kn^2)\) time complexity, where n is the number of SPD matrices we are taking the mean of, and k is their dimension, while the Euclidean mean for a set of covariances has a time complexity of \textit{O}\((n)\). Additionally, the need to replicate this alignment process for each class label is another efficiency bottleneck, which could be problematic in multi-class BCI paradigms. Introducing an intermediary dimensionality reduction step could expedite the computation of Riemannian Means, and adapting the method to be unsupervised might enhance efficiency for online applications. However, these modifications could compromise the enhancements seen over standard CSP in limited training data situations.

If a CSP-based spatial filter for a target subject performs poorly due to external factors such as suboptimal signal quality, diminished user attention, or motor-imagery illiteracy, our transfer paradigm is unlikely to significantly enhance the quality of the spatial filter. Essentially, the spatial filters that are aligned with a target subject will also inherit the characteristics of that target filter, including those contributing to its subpar performance. However, if the poor performance is primarily due to insufficient data, this is where our method can substantially boost performance, leveraging the transfer paradigm to compensate for data scarcity.

\section{Conclusion}

We have effectively demonstrated our method that combines two paradigms: Alignment of Covariance Matrices on Riemannian Manifolds and the direct use of Transfer Learning in CSP. As shown in this work and many preceding studies, the data shift between the covariances of a source and target subject can be modeled sufficiently with a composition of linear transformations, suggesting a potential role for contrastive learning in extracting representations of covariances that are invariant to these transformations. Integrating neural networks that respect the intrinsic geometry of SPD matrices within this framework, particularly in the context of contrastive learning on manifolds, represents a promising avenue for future work. 

\section{Appendix}

\subsection{Experimental Details}

We dedicate this section to explain the experimental details associated with figure 4. An explanation will be provided for one run of the experiment, with one condition, as in, what percentage of the target subjects training data was used, with the understanding that this procedure is simply repeated to obtain a distribution of the statistic we are interested in: how well RTCSP-SSF generalizes to unseen data compared to standard CSP When different amounts of data are available.

Suppose we have subjects \(\left\{S_1,S_2, \dots\ ,S_n\right\}\). For simplicity sake assume that each subject \(S_i\) has an associated dataset with shape \(N\times C \times T\), that is, N blocks of EEG that represent a motor imagery execution, with C channels, and T samples per channel. Additionally, suppose that we have some target subject \(S_k\), and we are interested in the condition where we use only p\%\ of the available training data from subject \(S_k\), thus, a total of \(\left\lceil{\frac{p}{100}N}\right\rceil\) motor imagery trials to obtain a spatial filter with. If \(S_k\) is the target subject then the source subjects are \(\left\{S_1,S_2, \dots\ ,S_n\right\} \setminus \left\{S_k\right\} \). We do not limit the training data on the remaining \(n - 1\) source subjects.

To compare standard CSP in low training data scenarios with RTCSP-SSF, we obtain two spatial filters. One using standard CSP with the limited \(\left\lceil{\frac{p}{100}N}\right\rceil\) training motor imagery trials, and we call this spatial filter \(W_{base}\). The other spatial filter results from the RTCSP-SSF method and will use \(\left\lceil{\frac{p}{100}N}\right\rceil + N(n-1)\) total motor imagery trials to obtain a spatial filter with. This spatial filter will be referred to as \(W_{RT}\). Suppose now that the target subject \(S_k\) has \(N_{te}\) test trials recorded from another session that were not used at all when obtaining \(W_{base}\) and \(W_{RT}\). We are interested in how well \(W_{base}\) and \(W_{RT}\) are able to generalize to these unseen trials and quantifying this for hypothesis testing purposes. 

We label these unseen test trials as \(X^{(1)}, \dots\ ,X^{(N_{te})}\) and map each one to the CSP space using \(W_{base}\) and \(W_{RT}\). For simplicity purposes consider a generic CSP-based spatial filter \(W\). We can map \(X^{(i)}\) to the CSP space with the transform \(X^{(i)}_{CSP} =  W^{'}X^{(i)}\). In this experiment we used one pair of spatial filters, thus \(X^{(i)}_{CSP} \) has two channels. Suppose that the variance of these two channels are \(v_1^{(i)}\) and \(v_2^{(i)}\), then the statistic we are interested in is
\[
\frac{\max(v_1^{(i)},v_2^{(i)})}{\min(v_1^{(i)},v_2^{(i)})}
\], which is directly related to the CSP objective function.

Using this statistic, we then compute the "mean-variance-ratio" (MVR) across all the test trials \(X^{(1)}, \dots\ ,X^{(N_{te})}\) 

\[
    MVR(W,S_k) = \frac{1}{N_{te}}\sum_{i = 1}^{N_{te}} \frac{\max(v_j^{(i)})}{\min(v_j^{(i)})}, j = 1,2
\]

The \(S_k\) in the argument of MVR referring to the fact that the above quantity is being calculated for test trials of subject \(S_k\) , and the W referring to what spatial filter is being used to project test trials into CSP space. This quantity should be obtained for all subjects, alternating between which subject is the target, which will also change which subjects are used as sources. Once we have this quantity for all subjects in a dataset, we simply average it to obtain our test statistic for one run of the experiment using the spatial filter \(W\). For subjects \(S_1, \dots , S_n\) our statistic for a single spatial filter W can be formulated as \(
MVR(W) = \frac{1}{n}\sum_{i=1}^{n} MVR(W,S_i)
\).

The quantity should be computed for both \(W = W_{base}\) (standard CSP) and \(W = W_{RT}\) (RTCSP-SSF). Due to the variability of this quantity, especially when less training data is used, this experiment was ran 50 times for each value of p, and during hypothesis testing we compare the distribution of \(MVR(W_{base})\) and \(MVR(W_{RT})\), with the alternative hypothesis that \(MVR(W_{RT}) > MVR(W_{base})\), meaning that RTCSP-SSF is able to map to a feature space where the resulting variance based features will be more discriminative on unseen data, 
which translates to more effective classification. We tested this hypothesis across nine different percentages of training data to use from a target subject, thus to conservatively account for multiple comparisons we set our alpha level to \(0.05/9\).

\begin{figure}[H]
\centering
\includegraphics[width = 0.9\textwidth]{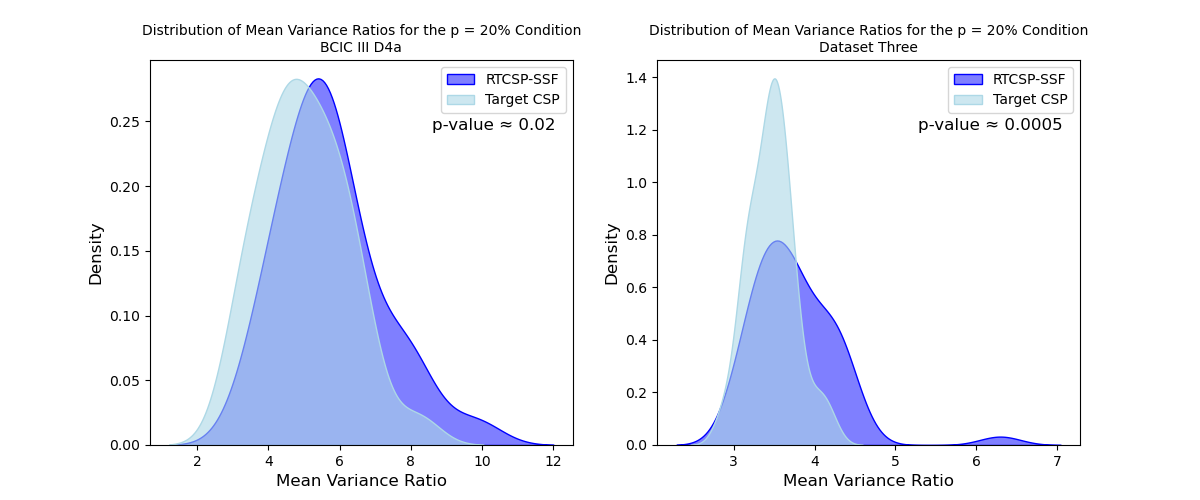}

\caption{{Example of 50 runs of the described experiment when 20\% of target training data is used. These distributions correspond with the left most bars on figure 4.}}

\end{figure}

\bibliographystyle{abbrv}
\bibliography{references}

\end{document}